\documentclass[letterpaper,10pt,conference]{ieeeconf}

\IEEEoverridecommandlockouts
\usepackage{graphicx}
\usepackage{amsmath}
\usepackage{amssymb}

\usepackage{url}
\usepackage[hidelinks]{hyperref}
\usepackage{xcolor}
\usepackage{xspace}
\usepackage{orcidlink}

\newtheorem{definition}{Definition}

\usepackage[xindy]{glossaries}

\usepackage{xspace}

\newcommand{\Varanus}{\textsc{Varanus}\xspace}

\newcommand{\assert}{\textit{ASSERT}$^{\text{TM}}$}

\newacronym{ltl}{LTL}{Linear-time Temporal Logic}

\newacronym{foltl}{FOLTL}{First-Order Linear-time Temporal Logic}

\newacronym{ctl}{CTL}{Computation Tree Logic}

\newacronym{ptl}{PTL}{Probabilistic Temporal Logic}

\newacronym{pctl}{PCTL}{Probabilistic Computation Tree Logic}

\newacronym{pctl*}{PCTL*}{Probabilistic Computation Tree Logic*}

\newacronym{tctl}{TCTL}{Timed Computation Tree Logic}

\newacronym{stl}{STL}{Signal Temporal Logic}

\newacronym{ltl-x}{LTL-X}{a version of Linear Time Logic without the `next' operator}

\newacronym{mtl}{MTL}{Metric Temporal Logic}

\newacronym{iactlk-x}{IACTLK-X}{a fragment of the temporal-epistemic logic CTLK, without the `next' operator}

\newacronym{mucalc}{$\mu$-calculus}{a strict superset of other temporal logics}

\newacronym{fotl}{FOTL}{First-Order Temporal Logic}

\newacronym{bdi}{BDI}{Belief-Desire-Intention}

\newacronym{prs}{PRS}{Procedural Reasoning System}

\newacronym{are}{ARE}{Autonomous Reasoning Engine}

\newacronym{dmars}{dMARS}{distributed Multi-Agent Reasoning System}

\newacronym{mas}{MAS}{Multi-Agent System}

\newacronym{asm}{ASM}{Abstract State Machines}

\newacronym{tasm}{TASM}{Timed Abstract State Machines}

\newacronym{fsm}{FSM}{Finite-State Machine}

\newacronym{pfsm}{PFSM}{Probabilistic Finite-State Machine}

\newacronym{apfsm}{APFSM}{\textit{Autonomous} Probabilistic Finite-State Machine}

\newacronym{pha}{PHA}{Probabilistic Hybrid Automata}

\newacronym{fsa}{FSA}{Finite-State Automata}

\newacronym{ta}{TA}{Timed Automata}

\newacronym{pta}{PTA}{Probabilistic Timed Automata}

\newacronym{gspn}{GSPN}{Generalised Stochastic Petri Net}

\newacronym{mopn}{MOPN}{Marked Ordinary Petri Net}

\newacronym{cpn}{CPN}{Coloured Petri Net}

\newacronym{ba}{BA}{B\"{u}chi Automata}

\newacronym{dtmc}{DTMC}{Discrete-Time Markov Chain}

\newacronym{ctmc}{CTMC}{Continuous-Time Markov Chain}

\newacronym{mdp}{MDP}{Markov Decision Process}

\newacronym{ioa}{IOA}{Input/Output Automata}

\newacronym{pars}{PARS}{Process Algebra for Robot Schemas}

\newacronym{fsp}{FSP}{Finite State Processes}

\newacronym{piadl}{$\pi$ADL}{$\pi$ calculus combined with ADL}

\newacronym{csp}{CSP}{Communicating Sequential Processes}

\newacronym{hcsp}{HCSP}{Hybrid Communicating Sequential Processes}

\newacronym{shcsp}{SHCSP}{Stochastic Hybrid Communicating Sequential Processes}

\newacronym{ccs}{CCS}{Calculus of Communicating Systems}

\newacronym{wsccs}{WSCCS}{Weighted Synchronous CCS}

\newacronym{csp|b}{CSP$\|$B}{an integrated formalism combining CSP and B}

\newacronym{vhdl}{VHDL}{VHSIC Hardware Description Language}

\newacronym{adl}{ADL}{Architecture Description Language}

\newacronym{slim}{SLIM}{System Level Integration Modelling}

\newacronym{mal}{MAL}{Model Action Logic Language}

\newacronym{acsl}{ACSL}{the ANSI C Specification Language}

\newacronym{fdr}{FDR}{the Failures-Divergences Refinement checker}

\newacronym{jpf}{JPF}{Java PathFinder}

\newacronym{ajpf}{AJPF}{Agent Java PathFinder}

\newacronym{mcmas}{MCMAS}{Model Checker for Multi-Agent Systems}

\newacronym{ltlmop}{LTLMoP}{Linear Temporal Logic MissiOn Planning}

\newacronym{cbmc}{CBMC}{C Bounded Model Checker}

\newacronym{cadp}{CADP}{Construction and Analysis of Distributed Processes}
\newacronym{spear}{SpeAR}{Specification and Analysis for Requirements Tool}

\newacronym{psl}{PSL}{Property Specification Language}

\newacronym{fpga}{FPGA}{Field-Programmable Gate Array}

\newacronym{aadl}{AADL}{Architecture Analysis and Design Language}

\newacronym{lts}{LTS}{Labelled-Transition System}

\newacronym{uml}{UML}{Unified Modelling Language}

\newacronym{sysml}{SySML}{Systems Modelling Language}

\newacronym{robotml}{RobotML}{Robot Modelling Language}

\newacronym{dsl}{DSL}{Domain Specific Language}

\newacronym{sct}{SCT}{Supervisory Control Theory}

\newacronym{sat}{SAT}{Boolean satisfiability problem}

\newacronym{smt}{SMT}{Satisfiability Modulo Theory}

\newacronym{ide}{IDE}{Integrated Development Environment}

\newacronym{mde}{MDE}{Model-Driven Engineering}

\newacronym{ai}{AI}{Artificial Intelligence}

\newacronym{fdir}{FDIR}{Fault Detection, Isolation and Recovery}

\newacronym{ifm}{iFM}{Integrated Formal Methods}

\newacronym{ros}{ROS}{Robot Operating System}

\newacronym{amcl}{AMCL}{Adaptive Monte Carlo Localisation}

\newacronym{dl}{\text{\upshape\textsf{d{\kern-0.05em}L}}}{differential dynamic logic}

\newacronym{vm}{VM}{Virtual Machine}

\newacronym{bip}{BIP}{Behaviour Interaction Priority}

\newacronym{fol}{FOL}{First-Order Logic}

\newacronym{owl}{OWL}{Web Ontology Language}

\newacronym{ria}{RIA}{Restore Invariant Approach}

\newacronym{ocl}{OCL}{Object Constraint Language}

\newacronym{sil}{SIL}{Safety Integrity Level}

\newacronym{rcl}{RCL}{ROS Contract Language}

\newacronym{rv}{RV}{Runtime Verification}

\newacronym{prv}{PRV}{Predictive Runtime Verification}

\newacronym{rml}{RML}{Runtime Monitoring Language}

\newacronym{sats}{SATS}{Small Aircraft Transportation System}

\newacronym{ico}{ICO}{Interactive Cooperative Objects}

\newacronym{gui}{GUI}{Graphical User Interface}

\newacronym{can}{CAN}{Controller Area Network}

\newacronym{pals}{PALS}{Physically Asynchronous, Logically Synchronous}

\newacronym{cps}{CPS}{Cyber-Physical System}

\newacronym{cots}{COTS}{Commercial Off-The-Shelf}

\newacronym{mtbf}{MTBF}{Mean Time Between Failures}

\newacronym{wcet}{WCET}{Worst-Case Execution Time}

\newacronym{sua}{SUA}{System Under Analysis}

\newacronym{caa}{CAA}{Civilian Aviation Authority}

\newacronym{atc}{ATC}{Air Traffic Control}

\newacronym{dlr}{DLR}{German Aerospace Centre}

\newacronym{assert}{\assert}{Analysis of Semantic Specifications and Efficient generation of Requirements-based Tests}

\newacronym{rae}{RAE}{Requirements Analysis Engine}

\newacronym{esa}{ESA}{European Space Agency}

\newcommand{\PVaranus}{\textsc{Predictive Varanus}\xspace}
\newcommand{\Lang}{\mathcal{L}\xspace}

\newcommand{\proj}{\pi\xspace}
\newcommand{\Buchi}{\mathcal{B}\xspace}
\newcommand{\AP}{\mathit{AP}\xspace}
\newcommand{\Cont}{\mathsf{Cont}\xspace}
\newcommand{\Pred}{\mathsf{Pred}\xspace}
\newcommand{\TS}{\mathcal{T}\xspace}

\title{\LARGE \bf
Predictive Varanus: Combining CSP Conformance Monitoring with Predictive LTL Runtime Verification
}

\author{Angelo Ferrando$^{1}$\orcidlink{0000-0002-8711-4670}, Matt Luckcuck$^{2}$\orcidlink{0000-0002-6444-9312}, and Pedro Ribeiro$^{3}$\orcidlink{0000-0003-4319-4872}%
\thanks{$^{1}$Angelo Ferrando is with the Department of Physics, Informatics and Mathematics, University of Modena and Reggio Emilia, Italy
        {\tt\small angelo.ferrando@unimore.it}}%
\thanks{$^{2}$Matt Luckcuck is with the School of Computer Science, University of Nottingham, UK {\tt\small matt.luckcuck@nottingham.ac.uk}
}
\thanks{$^{3}$Pedro Ribeiro is with the Department of Computer Science, University of York, UK {\tt\small pedro.ribeiro@york.ac.uk}
}
}

\begin{document}

\maketitle
\thispagestyle{empty}
\pagestyle{empty}

\begin{abstract}
Runtime Verification is well suited to autonomous and robotic systems because it checks the behaviour that is actually observed during execution. Its main limitation, however, is that it is usually reactive: the monitor detects a violation only after the system has already performed a bad event. This can be too late in domains where failures are costly or unsafe. In this paper we present \PVaranus, a two-stage verification pipeline that combines \Varanus, a runtime verifier that uses models written in the process algebra \gls{csp}, with predictive runtime verification for LTL. A \gls{csp} model is first used as a conformance gate over the observed event trace; the same model is then translated into a B\"uchi automaton that constrains the futures explored by a predictive LTL monitor. In this way, out-of-model behaviour is rejected immediately, while model-consistent prefixes can be classified as already guaranteeing satisfaction, already forcing violation, or still being inconclusive for the monitored temporal property. We formalise the combined monitor, explain its implementation, and illustrate the approach on a robotic rover for nuclear-store inspection. The case study shows how the combination of \gls{csp} validation and predictive LTL can provide earlier verdicts than standard runtime monitoring while reusing an existing design-time \gls{csp} model.
\end{abstract}
\glsresetall

\section{Introduction}

Autonomous robotic systems often operate in environments that are only partially known at design time. For this reason, offline analysis and design-time verification alone are often insufficient: assumptions that held in the design model may be violated at runtime, sensors may be exposed to unexpected situations, and interactions with the environment can steer the system into untested behaviours. \gls{rv} addresses this gap by checking the trace of events produced by the \gls{sua} against a formal specification of its intended behaviour \cite{DBLP:conf/kbse/GiannakopoulouH01,DBLP:journals/tosem/BauerLS11}, this formal specification is know as an \textit{oracle}. In robotics, \gls{rv} is particularly useful because it can provide online assurance that the robot is obeying its specification while running, without needing access to the source code or the internal state of the system's components~\cite{DBLP:journals/csur/LuckcuckFDDF19}.

A limitation of standard \gls{rv} is that it is reactive: unless it is coupled with enforcement or mitigation, the monitor reports violations only after the relevant event has been observed. \gls{prv} addresses a complementary problem by using a model of the \gls{sua} to look ahead at the continuations that remain possible after the current trace, and to check whether a property will now inevitably hold or inevitably fail \cite{DBLP:conf/rv/Leucker12,DBLP:conf/nfm/ZhangLD12,DBLP:journals/jss/PinisettyJTFMP17}. These verdicts are conditional on the model: they do not predict arbitrary unmodelled failures, but anticipate model-consistent outcomes such as inevitable mission abort, failed coverage, or guaranteed completion.

Our recent work introduces \Varanus~\cite{DBLP:conf/taros/LuckcuckFF25}, an \gls{rv} tool that uses a \gls{csp} \cite{DBLP:series/txcs/Roscoe10} model as its oracle. \Varanus assumes that the \gls{csp} model is a deterministic representation of the correct behaviour; but otherwise requires no modifications, which means that we can reuse a model that has already been analysed during design, for example using FDR~\cite{DBLP:conf/tacas/Gibson-RobinsonABR14} or RoboChart~\cite{DBLP:journals/sosym/MiyazawaRLCTW19}. This is a useful approach for monitoring autonomous systems, because \gls{csp} naturally captures event ordering, synchronisation, and communication structure, and because many robotic specifications are already written at that level of abstraction. However, \Varanus is a reactive monitor not a predictive monitor, it can only conclude that the \gls{sua} is misbehaving when an invalid event is observed.


This paper combines our previous work on \Varanus with \gls{prv}, presenting a pipeline called \PVaranus. The pipeline uses \Varanus as a \gls{csp} conformance gate, and then applies \gls{prv} of \gls{ltl} properties to the futures that remain compatible with the current trace. Our pipeline can detect two different situations. First, \gls{sua} behaviour that does not conform with the \gls{csp} model, which will be rejected. Second, \gls{sua} behaviour that is conformant with the model, but from which we can conclude the inevitable satisfaction or violation of an \gls{ltl} property.

Our pipeline demonstrates how to achieve \gls{prv} of temporal logic from a \gls{csp} model. Here, we use the \gls{csp} model in two ways. First, as the oracle for \Varanus when used as a conformance gate. Second, as the source of admissible futures for predictive reasoning, for which we translate the \gls{csp} model into a B\"uchi automaton and project the \gls{sua} event to propositions.
In this paper, the \gls{prv} is implemented as \gls{ltl} checking, but the pipeline is modular so other predictive reasoning could be used as needed.
%
%
We define the resulting two-stage pipeline, formalise its combined verdict function and model translation, and evaluate it on a nuclear-inspection rover case study.

The rest of the paper is organised as follows. Section~II describes background information about \Varanus and \gls{prv}. Section~III presents the \PVaranus pipeline and its formalisation. Section~IV applies \PVaranus to our case study. Section~V presents the evaluation.
Section~VI reviews related work. Section~VII concludes.

\section{Background}

\subsection{CSP and \Varanus}

\gls{csp} models systems as processes that engage in events. 
\Varanus assumes its \gls{csp} oracle is a deterministic, finite-state process.
\gls{csp}'s operational semantics can be represented as a \gls{lts}, in which states correspond to process configurations and edges are labelled by visible events. The traces of a process are the finite paths through the \gls{lts}. This
makes \gls{csp} particularly well suited to \gls{rv} because
the \gls{sua}'s execution is a trace of events that can be  compared to the traces of the model.

\Varanus uses the \gls{lts} representation of a \gls{csp} model as its \gls{rv} oracle. In the original \Varanus workflow, an event is accepted if it is available in the current state of the \gls{lts}, and rejected otherwise. \Varanus has two modes: strict and permissive. Permissive mode ignores events that are outside the oracle's alphabet, which is useful when the \gls{sua} emits irrelevant events. 
In this paper, \Varanus plays the role of a \emph{conformance gate}: only events that are consistent with the \gls{csp} model are passed to the predictive component.

For the remainder of the paper, let $P$ be a \gls{csp} process over the visible
alphabet $\Sigma_P$. We write $\Lang(P)$ for the set of finite traces admitted
by $P$. Since the trace semantics of \gls{csp} is prefix-closed
\cite{DBLP:series/txcs/Roscoe10}, $\sigma \in \Lang(P)$ means that the observed
finite trace $\sigma$ is still compatible with the \gls{csp} specification.

\begin{definition}[CSP conformance predicate]
Given a \gls{csp} process $P$, the conformance predicate is the function
\[
G_P : \Sigma_P^* \rightarrow \{\mathsf{acc},\mathsf{rej}\}
\]
defined by
\[
G_P(\sigma)=
\begin{cases}
\mathsf{acc} & \text{if } \sigma \in \Lang(P),\\
\mathsf{rej} & \text{otherwise.}
\end{cases}
\]
\end{definition}

This definition captures \Varanus's role in the pipeline:
\Varanus evaluates whether the currently observed trace is a valid trace of the \gls{csp} model. If this is the case, $G_P = acc$, then the trace is checked by the predictive component. But if the trace is rejected, $G_P = rej$, then the \gls{sua} has performed an invalid event so the predictive component is not needed.

\subsection{Predictive Runtime Verification for LTL}

Let $\varphi$ be an \gls{ltl} property over an alphabet $\Sigma$, and let $M$ be a model describing the system's executions over the same alphabet. In \gls{prv}, the verdict on a finite prefix $\sigma$ depends not only on whether $\sigma$ satisfies or violates the property $\varphi$, but on whether every continuation of $\sigma$ allowed by $M$ will satisfy it, every continuation will violate it, or both are possible~\cite{DBLP:conf/nfm/ZhangLD12,DBLP:journals/jss/PinisettyJTFMP17,DBLP:journals/fmsd/FerrandoCFLPFM21}.

For this paper we adopt a three-valued logic that suits our implementation: conclusive satisfaction ($\top$), conclusive violation ($\bot$), and inconclusive ($?$). This is a collapsed version of richer predictive semantics, where the intermediate ``possibly true'' and ``possibly false'' verdicts are merged into a single inconclusive result, because \Varanus will reject invalid prefixes.

\begin{definition}[Predictive verdict]
Let $M$ be a model over $\Sigma$, let $\varphi$ be an \gls{ltl} property, and let
\[
\Cont_M(\sigma)=\{\tau \in \Sigma^\omega \mid \sigma\tau \in \Lang(M)\}.
\]
Then, for every finite trace $\sigma \in \Sigma^*$,
\[
\Pred_{\varphi,M}(\sigma)\in\{\top,\bot,?\}
\]
is defined by
\[
\Pred_{\varphi,M}(\sigma)=
\begin{cases}
\top & \text{if } \forall \tau \in \Cont_M(\sigma).\ \sigma\tau \models \varphi,\\
\bot & \text{if } \forall \tau \in \Cont_M(\sigma).\ \sigma\tau \not\models \varphi,\\
?    & \text{otherwise.}
\end{cases}
\]
\end{definition}

To enable us to use standard \gls{ltl} semantics, we extend a finite trace with an infinite suffix of \texttt{skip} propositions. Events in the trace produced by the \gls{sua} satisfy $\neg \texttt{skip}$, with an infinite suffix of \texttt{skip} added after the trace terminates. In \gls{csp} this suffix would correspond to an infinite continuation of \texttt{skip} events. 


\section{Predictive Varanus}

\begin{figure*}[t]
    \centering
    \includegraphics[width=0.5\linewidth]{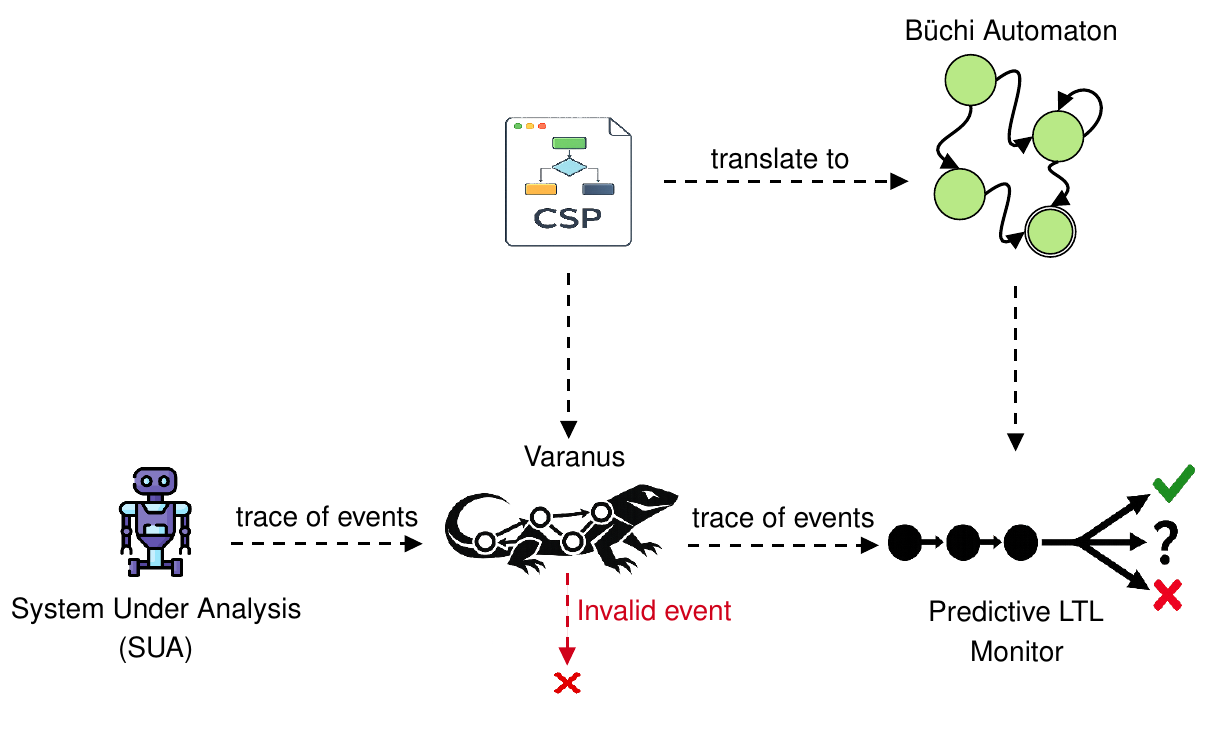}
    \caption{The \PVaranus verification pipeline. A \gls{csp} model is reused both as a runtime conformance gate through \Varanus and as the source of the B\"uchi automaton used by the predictive LTL monitor.}
    \label{fig:pipeline}
\end{figure*}

Figure~\ref{fig:pipeline} shows the components in the \PVaranus pipeline. The \gls{sua} emits a trace of events. Each event is checked by \Varanus for validity with the \gls{csp} model, if it is a valid event then it is passed to the Predictive \gls{ltl} Monitor.
The predictive monitor projects the event into the propositional alphabet, and uses the B\"uchi automaton (translated from the \gls{csp} model) to perform \gls{prv}.
In this way, the predictive component never reasons about futures for events that \Varanus has already rejected as invalid.

\subsection{From \gls{csp} to a proposition-based model}

A translation from \gls{csp} models to proposition-level runtime
verification already exists, notably through translations of \gls{csp} normalised
graphs to Kripke structures for sound \gls{rv} of traces and
refusals~\cite{DBLP:conf/ictac/CavalcantiHPW15}. Our construction is different in purpose: rather than
targeting safety, we derive a propositional
$\omega$-automaton used for predictive \gls{ltl} monitoring over the futures for events
accepted by the \gls{csp} model.

\paragraph{From \gls{csp} to the predictive model.}
Let $P$ be a finite-state \gls{csp} process. The first step is to
extract from $P$ the finite \gls{lts} induced by its standard
operational semantics. Formally, we write
\[
\mathcal{T}(P)=(S,s_0,\Sigma_P,\rightarrow),
\]
where $S$ is the set of reachable residual states of $P$, $s_0 \in S$ is the
initial state, $\Sigma_P$ is the event alphabet of $P$, and
\[
\rightarrow \;\subseteq S \times \Sigma_P \times S
\]
is the transition relation. For $s,s' \in S$ and $a \in \Sigma_P$, the
transition
\[
s \xrightarrow{a} s'
\]
holds whenever, according to the standard \gls{csp} operational semantics, the
process can evolve from state $s$ to state $s'$ by performing event $a$.

Intuitively, each state of $\mathcal{T}(P)$ represents the residual behaviour
of the process after some finite execution, and each labelled transition
records a possible next event. Since $P$ is finite-state, the transition system
is finite as well. In the implementation, \Varanus obtains this structure by
delegating \gls{csp} parsing and state-space generation to FDR, and then materialises
the reachable transition system explicitly for runtime use. Thus, the
transition system employed by the monitor is not an ad hoc artefact of the
implementation, but the explicit finite-state representation of the standard
semantics of the \gls{csp} model.

To interface the \gls{csp} model with the predictive \gls{ltl} layer, we derive
from $\TS(P)$ an $\omega$-automaton encoded in the Hanoi Omega-Automata (HOA)
format
\[
\Buchi_P=(S,s_0,2^{\AP},\delta,F_P),
\]
where
\[
\AP=\{\proj(a)\mid a\in\Sigma_P\}\cup\{\texttt{skip}\}
\]
and $F_P = S$, that is, every state is accepting. This is analogous to the
standard use of Kripke structures in automata-based temporal verification:
\gls{ltl} is interpreted over infinite behaviours, so terminal states must be
completed with an infinite stuttering suffix. In our setting, this is achieved
by adding a \texttt{skip} self-loop to every terminal state. As a result, every
maximal finite execution of the \gls{csp} model induces an infinite word, and
$\Buchi_P$ accepts exactly those infinite continuations obtained from the
behaviour of the \gls{csp} model under this terminal-stuttering convention. The
mapping $\proj$ associates each visible \gls{csp} event with a fresh atomic
proposition, while \texttt{skip} is reserved for the post-termination suffix.

For every \gls{csp} transition
\[
s \xrightarrow{a} s',
\]
the automaton contains the transition
\[
s \xrightarrow{\lambda(a)} s',
\]
where $\lambda(a)\in 2^{\AP}$ is the one-hot valuation associated with $a$,
namely
\[
\lambda(a)=\{\proj(a)\}.
\]
Equivalently, in Boolean form, this label asserts $\proj(a)$, negates all other
event propositions, and also negates \texttt{skip}.

If a state $s \in S$ is terminal, that is,
\[
\forall a\in\Sigma_P.\ \nexists s' \in S \text{ such that } s\xrightarrow{a}s',
\]
then we add the post-termination suffix
\[
s \xrightarrow{\{\texttt{skip}\}} s.
\]
This turns every maximal finite execution into an infinite word, so that
standard \gls{ltl} semantics can be used.

Although \Varanus produces a HOA-formatted automata, the predictive monitor uses it as a transition system: every infinite run induced by
$\delta$ is treated as an admissible continuation of the current prefix.
Hence, after observing a finite trace $\sigma$, the continuations considered by the predictive monitor are precisely the infinite propositional words that can still be generated from the state reached in $\Buchi_P$ after reading $\sigma$.

The automaton $\Buchi_P$ derived from the \gls{csp} model provides the set of
behaviours that are still compatible with the current execution prefix. After a 
finite trace $\sigma$ has been read, the predictive monitor considers all
infinite runs of $\Buchi_P$ that may continue from the state reached after
$\sigma$. These runs represent the futures still admitted by the \gls{csp} model and
are used to determine whether the monitored \gls{ltl} property is necessarily
satisfied, necessarily violated, or still undetermined.

This projection is not merely a technical encoding. In \gls{csp}, events can have a richer structure than is directly expressible in propositional \gls{ltl}. Events can be parameterised, for example
$\mathit{move}.i$, with $i \in \{0,\ldots,N\}$, to denote that the robot is
commanded to move to waypoint $i$. The concrete events $\mathit{move}.0$, $\mathit{move}.1$, \ldots, $\mathit{move}.N$ are therefore distinct at the \gls{csp} level, and in our
construction they are mapped to distinct atomic propositions, for example
$\proj(\mathit{move}.0)=m_0$ and $\proj(\mathit{move}.1)=m_1$. Thus, the
predictive monitor does not reason over the structured \gls{csp} events
directly, but it still distinguishes their propositional images. What is lost
in the projection is not the distinction between these events, but their
original structured representation as parameterised \gls{csp} events.
While \Varanus validates the observed event against the \gls{csp} model using
its original event structure, the predictive monitor consumes only the
corresponding propositional encoding.

This projection is important for two reasons. First, it lets us reuse mature
automata-based \gls{ltl} tooling without needing to change the \gls{csp} model. Second, it cleanly separates the structural validation of
runtime events, which is done by \Varanus; from the predictive monitoring, which is performed on the abstract propositional trace.

\subsection{Predictive monitoring construction}

Given an \gls{ltl} formula $\varphi$, let $\Buchi_{\varphi}$ and
$\Buchi_{\neg\varphi}$ be the B\"uchi automata for $\varphi$ and $\neg\varphi$,
respectively. The predictive monitor is based on the two synchronous products
\[
A^{+} = \Buchi_P \times \Buchi_{\varphi},
\qquad
A^{-} = \Buchi_P \times \Buchi_{\neg\varphi}.
\]
For a finite trace $\sigma$, let $S^{+}_{\sigma}$ and $S^{-}_{\sigma}$ denote
the sets of states reachable in $A^{+}$ and $A^{-}$, respectively, after
reading the projected trace $\proj(\sigma)$. Intuitively, $S^{+}_{\sigma}$
captures the model-consistent futures that still admit satisfaction of
$\varphi$, while $S^{-}_{\sigma}$ captures those that still admit violation of
$\varphi$.

\begin{definition}[Predictive monitor verdict]
For a finite trace $\sigma$, the predictive monitor returns
\[
R_{P,\varphi}(\sigma)=
\begin{cases}
\top & \text{if } \Lang(A^{-},S^{-}_{\sigma})=\emptyset,\\
\bot & \text{if } \Lang(A^{+},S^{+}_{\sigma})=\emptyset,\\
? & \text{otherwise,}
\end{cases}
\]
where $\Lang(A,S)$ denotes the language accepted by automaton $A$ when
initialised in the state set $S$.
\end{definition}

This verdict criterion is not specific to our framework, but follows the
standard automata-theoretic formulation of predictive runtime verification
\cite{DBLP:conf/nfm/ZhangLD12}. The key question is whether, after the current
finite trace, there still exists an accepting continuation in the product with
$\Buchi_{\varphi}$ and in the product with $\Buchi_{\neg\varphi}$. If the
former language is empty, then violation is inevitable; if the latter is empty,
then satisfaction is inevitable; otherwise, both outcomes remain compatible
with the current execution and the verdict stays inconclusive. Accordingly, our
contribution is not a new predictive verdict rule, but the integration of that
standard rule with a \gls{csp}-based behavioural model: the admissible futures are
not postulated externally, but extracted from the \gls{csp} specification and
continuously validated online by \Varanus according to the actually observed
execution.

\subsection{Combined semantics}

The full \PVaranus pipeline is the sequential composition of the \gls{csp}
conformance check and the predictive \gls{ltl} monitor.

\begin{definition}[Predictive Varanus]
Given a \gls{csp} model $P$ and an \gls{ltl} property $\varphi$, the combined monitor on a
finite trace $\sigma$ is defined as
\[
\mathsf{PV}_{P,\varphi}(\sigma)\in\{\mathsf{rej},\top,\bot,?\}
\]
by
\[
\mathsf{PV}_{P,\varphi}(\sigma)=
\begin{cases}
\mathsf{rej} & \text{if } G_P(\sigma)=\mathsf{rej},\\
R_{P,\varphi}(\sigma) & \text{if } G_P(\sigma)=\mathsf{acc}.
\end{cases}
\]
\end{definition}

This distinction is semantically important. The verdict $\mathsf{rej}$ does not indicate a violation of the monitored \gls{ltl} property, but that the trace has left the behaviours admitted by the \gls{csp} model. In that case, the assumptions used for prediction no longer apply. By contrast, $\bot$ denotes a predictive violation of the \gls{ltl} property along a trace that is still model-conformant. \PVaranus therefore anticipates outcomes that are forced by the remaining model-consistent futures, such as mission abort, failed coverage, or guaranteed completion, rather than unmodelled physical or sensing failures.

\section{Case Study}
\label{sec:case-study}

To evaluate \PVaranus, we reuse the example of 
an autonomous rover patrolling a simulated nuclear waste store, and reuse the \gls{csp} model of the rover, that we used in our earlier \Varanus work~\cite{DBLP:conf/taros/LuckcuckFF25}. 
The rover's mission is to inspect the radiation levels at various waypoints. The radiation readings are categorised into \emph{Green}, \emph{Orange}, or \emph{Red}; if the radiation is \emph{Green} then the mission can continue, but \emph{Orange}, or \emph{Red} mean that the rover should return to the entry point and abort the mission. 
Using the same \gls{csp} model and mission lets us isolate the added value of predictive runtime verification over plain runtime validation.

In our \gls{csp} model, the rover repeatedly selects a waypoint, moves to it, and reacts to the radiation reading.
\Varanus can detect incorrect behaviour after detecting high radiation, such as moving to another waypoint instead of the entry point.
But our aim with \PVaranus is to anticipate situations where some mission requirements are impossible (or inevitable), if the rover obeys the model.

We use \PVaranus to monitor three properties: $\varphi_{\mathit{complete}}$, $\varphi_{\mathit{abort}}$, and $\varphi_{\mathit{cover}}$, defined below.
Let $mc$ abbreviate the proposition corresponding to \texttt{mission\_complete}, $ma$ the one for \texttt{mission\_abort}, $r$ and $o$ the propositions for \texttt{radiation\_level.Red} and \texttt{radiation\_level.Orange}, and $m_i$ the proposition for \texttt{move.$i$}.
The first property asks whether the mission can still be completed successfully:
\[
\varphi_{\mathit{complete}} = F(\neg skip \wedge mc).
\]
After the monitor has observed a prefix such as
\[
\langle \texttt{mission\_start},\texttt{inspect.2},
\]
\[
\texttt{radiation\_level.Red} \rangle,
\]
this property becomes predictively false, despite \texttt{mission\_abort} not having happened, because the \gls{csp} model shows that every continuation from that state returns to the entry point and aborts. Therefore, successful mission completion is no longer compatible with the model.

The second property captures the safety obligation that hazardous radiation eventually forces mission abortion:
\[
\varphi_{\mathit{abort}} = G((r \vee o) \rightarrow F(\neg skip \wedge ma)).
\]
For the same prefix as above, this property becomes predictively true before
\texttt{mission\_abort} is observed, because every model-consistent future now
contains a \texttt{mission\_abort} event.

A third property captures full mission coverage. If the mission is expected to visit all five inspection waypoints in the nominal case, we can write
\[
\varphi_{\mathit{cover}} = \bigwedge_{i=1}^{5} F(\neg skip \wedge m_i).
\]
While the mission progresses with low (\emph{Green}) radiation readings, this property is inconclusive ($?$). 
If \PVaranus detects that the model can now only end with a \texttt{mission\_abort} event, and there are unvisited waypoints, then the property becomes predictively false. This is a useful example because the trace prefix is still conformant with the model, yet the monitor can already tell that the coverage requirement is unattainable (assuming the rover obeys the model). This would be useful information for a human operator or for re-planning.

These examples illustrate why the combination of \gls{csp} and predictive \gls{ltl} is useful. The \gls{csp} model is used to both recognise bad events and predict future events. \PVaranus's verdict reuses the rover's design model to constrain the continuations it needs to check, rather checking every combination of events the rover can perform.

\section{Evaluation}

This section presents two evaluations of
\PVaranus\footnote{\PVaranus available at:
\url{https://github.com/AngeloFerrando/PredictiveVaranus}.}. First, we study
the rover mission from Sect.~\ref{sec:case-study} to validate the verdicts and
quantify the practical benefit of the approach. Second, we conduct a
quantitative stress test on synthetic \gls{csp} models to assess computational
overhead and scalability under controlled variations of model size, branching
structure, and termination depth.

We separate the cost of \Varanus from the additional cost and benefit of the
predictive monitor. Since \Varanus and \PVaranus do not solve the same
monitoring task, we do not compare them directly. Instead, we measure the
incremental cost of prediction on top of \gls{csp} validation, and the benefit
of prediction as the difference between the position at which \PVaranus returns
a conclusive verdict and the later position at which the same conclusion would
become explicit without prediction.

\subsection{Rover Evaluation}

Our goals for this evaluation are to check that the combined monitor yields the expected verdicts, and to measure how many events earlier than the invalid event that the predictive monitor can conclude a verdict.

\subsubsection{Qualitative Measures}

We consider the following three three trace classes.

\paragraph{Invalid traces}
Some traces violate the \gls{csp} model, e.g. the rover continues its mission after a hazardous radiation reading, or 
moves to a waypoint that it was not told to inspect.
In such cases, \Varanus rejects the trace at the first illegal event, showing that the predictive monitor does not alter the original conformance semantics.

\paragraph{Conformant traces with predictively false properties}
Other traces are conformant with the model, but the predictive monitor can conclude that one or more properties are unreachable. 
If hazardous radiation is detected, then the mission completion and coverage properties become predictively false before the \texttt{mission\_abort} event, assuming that the rover obeys the model. This is the clearest case of genuinely predictive negative verdicts.

\paragraph{Conformant traces with predictively true properties}
The third class consists of conformant traces where the satisfaction of a property has already become inevitable, again assuming that the rover obeys the model. 
If radiation is detected, then the abort property predictively true before the \texttt{mission\_abort} event happens.
Conversely, when all the waypoints have been inspected, the coverage property becomes predictively true, even though the rover will perform more events before the \texttt{mission\_complete} event.

In summary, \Varanus rejects non-conformant traces at the first invalid event, whereas the predictive monitor anticipates inevitable satisfaction/violation on conformant prefixes.

\subsubsection{Quantitative measures}

For each property and trace, we record:
(i) the first event position where \PVaranus returns a conclusive verdict;
(ii)  the later reference position where the same conclusion becomes explicit in the trace, and;
(iii) the anticipation gain, \textit{i.e.}\ the difference between the two.
The monitored rover properties are
$\varphi_\mathit{complete}$,
$\varphi_\mathit{cover}$, and
$\varphi_\mathit{abort}$.
Table~\ref{tab:rover-eval} reports our results over 27 separate runs.
\begin{table}[t]
\centering
\scriptsize
\caption{Results from evaluation on our rover case study. \emph{Pred.\ pos.} is the first conclusive
\PVaranus verdict; \emph{Ref.\ pos.} is the later position where the same
verdict happens.}
\label{tab:rover-eval}
\begin{tabular}{|p{0.25\columnwidth}|p{0.13\columnwidth}|p{0.09\columnwidth}|p{0.09\columnwidth}|p{0.09\columnwidth}|p{0.06\columnwidth}|}
\hline
\textbf{Scenario} & \textbf{Property} & \textbf{Verdict} & \textbf{Pred.\ pos.} & \textbf{Ref.\ pos.} & \textbf{Gain} \\
\hline
Nominal mission & $\varphi_{\mathit{complete}}$ & True  & 25 & 25 & 0 \\
\hline
Nominal mission & $\varphi_{\mathit{cover}}$    & True  & 17 & 25 & 8 \\
\hline
Red hazard then abort & $\varphi_{\mathit{complete}}$ & False & 9  & 11 & 2 \\
\hline
Red hazard then abort & $\varphi_{\mathit{abort}}$    & True  & 10 & 11 & 1 \\
\hline
Orange hazard then abort & $\varphi_{\mathit{complete}}$ & False & 13 & 15 & 2 \\
\hline
Invalid red hazard then continue & $\varphi_{\mathit{complete}}$ & False & 10 & 10 & 0 \\
\hline
\end{tabular}
\end{table}
As Table~\ref{tab:rover-eval} shows, prediction gives no benefit for
$\varphi_{\mathit{complete}}$ on the nominal trace, since completion can only
be established at \texttt{mission\_complete}, but yields an 8-event gain for
$\varphi_{\mathit{cover}}$, which becomes inevitable once the last waypoint is
visited. After hazardous radiation is detected,
$\varphi_{\mathit{complete}}$ and $\varphi_{\mathit{cover}}$ become
predictively false two events before \texttt{mission\_abort}, while
$\varphi_{\mathit{abort}}$ becomes predictively true one event earlier.
Invalid traces show zero gain, since \Varanus rejects them at the first illegal
action.

We observe the same behaviour in the other abort scenarios, where the rover
aborts after making further mission progress. In each case,
$\varphi_{\mathit{complete}}$ and $\varphi_{\mathit{cover}}$ become
conclusively false two events before the abort event, while
$\varphi_{\mathit{abort}}$ becomes conclusively true one event earlier.

Runtime overhead is modest across all 27 runs: the monitoring time is $1.84$ ms/event, of which \Varanus takes $0.19$ ms/event and the predictive monitor $1.64$ ms/event.
Per property, the average cost is $0.32$ ms/event for $\varphi_{\mathit{complete}}$, $0.35$ ms/event for
$\varphi_{\mathit{abort}}$, and $4.84$ ms/event for
$\varphi_{\mathit{cover}}$, consistent with the larger automata needed for
coverage.

\subsection{Stress-Test Evaluation}

We complement the rover evaluation with synthetic benchmarks over generated
\gls{csp} models and traces. The stress test has two parts. The first varies
the size of a densely connected \gls{csp} model in order to measure how the
runtime cost scales as the number of states and transitions increases. The
second uses traces with a branching decision prefix followed by a forced tail,
so as to measure predictive benefit when the final outcome becomes explicit only
after a controllable delay.

\subsubsection{Experimental setup}

Experiments were run on Ubuntu~24.04.4~LTS
(\texttt{Linux 6.8.0-106-generic}) on a server with an 11th Gen Intel Core
i9-11900KF CPU (3.50\,GHz, 8 cores, 16 threads) and 32\,GiB RAM. Benchmarks
were executed in Docker without CPU or memory limits. The predictive monitor
uses Python~3.8.10 and Spot~2.12; the original \Varanus backend uses
Python~2.7.18.

In the first synthetic benchmark, we vary the size of a densely connected CSP
model. Here, $N$ is the number of states in the generated model, and $L$ is the
length of the trace processed at runtime. 
%
In the second synthetic benchmark, traces consist of a branching decision
prefix followed by a forced tail. Here, $B$ is the branching factor of the
decision prefix, $D$ is its depth, and $T$ is the length of the forced tail
before the terminal event. 

\subsubsection{Cost Metrics}

For each run on the dense synthetic benchmark, we record a one-off
preprocessing cost and the average per-event monitoring cost. The preprocessing
cost includes constructing the propositional alphabet from the observable CSP
events, exporting the resulting model automaton in HOA format, rewriting the
monitored LTL formula over the same propositional alphabet, and initialising
the predictive monitor. The per-event cost is then decomposed into the cost of
the \Varanus conformance check and the additional cost of the predictive
monitor.
Table~\ref{tab:stress-test-costs} reports these results for the dense synthetic
benchmark. Here, $N$ is the number of states in the generated \gls{csp} model,
and \emph{Trace length} is the number of runtime events processed.
\emph{Preproc.\ time} reports the one-off preprocessing cost, and the remaining
columns report average per-event costs for \gls{csp} validation, predictive
monitoring, and their sum.

\begin{table*}[t]
\centering
\scriptsize
\caption{Stress-test cost decomposition on the dense synthetic family. Times
are in milliseconds. Preprocessing is a one-off cost; the remaining columns are
average per-event costs during online monitoring.}
\label{tab:stress-test-costs}
\begin{tabular}{|c|c|c|c|c|c|}
\hline
\textbf{Model parameter} & \textbf{Trace length} & \textbf{Preproc.\ time} & \textbf{CSP cost/event} & \textbf{Pred.\ cost/event} & \textbf{Total cost/event} \\
\hline
$N=10$  & $100$   & 95.37      & 0.129 & 0.198   & 0.327 \\
\hline
$N=10$  & $1000$  & 94.48      & 0.109 & 0.179   & 0.288 \\
\hline
$N=10$  & $10000$ & 95.31      & 0.100 & 0.176   & 0.276 \\
\hline
$N=50$  & $100$   & 1787.79    & 0.356 & 7.544   & 7.900 \\
\hline
$N=50$  & $1000$  & 1807.28    & 0.314 & 7.596   & 7.911 \\
\hline
$N=50$  & $10000$ & 1785.69    & 0.329 & 7.604   & 7.934 \\
\hline
$N=100$ & $100$   & 25207.81   & 0.577 & 64.462  & 65.039 \\
\hline
$N=100$ & $1000$  & 25120.47   & 0.582 & 65.102  & 65.685 \\
\hline
$N=100$ & $10000$ & 25106.34   & 0.598 & 65.482  & 66.082 \\
\hline
$N=200$ & $100$   & 389574.41  & 0.972 & 756.577 & 757.550 \\
\hline
$N=200$ & $1000$  & 391505.33  & 0.984 & 705.667 & 706.653 \\
\hline
$N=200$ & $10000$ & 396888.37  & 0.992 & 701.640 & 702.633 \\
\hline
\end{tabular}
\end{table*}

The cost results show a clear scaling pattern. The \gls{csp}-validation
backbone remains relatively cheap throughout, staying below $1$\,ms/event even
at $N=200$. By contrast, the predictive monitor dominates the online cost as
the model grows, increasing from about $0.2$\,ms/event at $N=10$ to about
$7.6$\,ms/event at $N=50$, $65$\,ms/event at $N=100$, and
$702$--$757$\,ms/event at $N=200$. For fixed $N$, per-event costs remain
fairly stable across trace lengths, indicating that online cost depends mainly
on automaton size rather than on the total length of the trace.

The one-off preprocessing cost shows the same trend even more strongly. It
grows from about $95$\,ms at $N=10$, to about $1.8$\,s at $N=50$, to about
$25$\,s at $N=100$, and to roughly $390$--$397$\,s at $N=200$. Thus, the main
scalability bottleneck of the current implementation is not the \gls{rv} phase that uses \Varanus, but the automata construction and maintenance required by the predictive
monitor in the \gls{prv} phase.

\subsubsection{Prediction Metrics}

For the second synthetic benchmark, the main metric is anticipation gain. Each
generated trace has four parts: an initial \emph{decision prefix}, in which the
execution may follow one of several branches; a single \emph{commitment event},
after which the eventual outcome is fixed; a \emph{forced tail}, namely a
sequence of events determined by that commitment; and a final \emph{terminal
event} that makes the outcome explicit. The predictive monitor is expected to
conclude at the commitment event, because from that point onward all remaining
model-consistent continuations agree on the verdict, whereas the reference
point is the final terminal event, where the same verdict becomes explicit
without prediction.
In total, the benchmark yields 96 aggregated configurations
($3$ branching factors $\times 2$ depths $\times 4$ tail lengths $\times 2$
properties $\times 2$ trace classes). The results are highly regular:
branching factor, property polarity, and trace class do not affect the measured
gain, while decision depth shifts predictive and reference latencies by the
same constant offset. Table~\ref{tab:stress-test-benefit} therefore reports
representative rows only.

\begin{table*}[t]
\centering
\scriptsize
\caption{Representative decision-tail predictive-benefit results. The omitted 96-row
table follows the same pattern: for fixed depth $D$, predictive latency is
constant, reference latency increases with tail length $T$, and gain depends
only on $T$.}
\label{tab:stress-test-benefit}
\begin{tabular}{|c|c|c|c|c|c|}
\hline
\textbf{Model parameter} & \textbf{Property} & \textbf{Trace class} & \textbf{Pred.\ latency} & \textbf{Ref.\ latency} & \textbf{Anticipation gain} \\
\hline
$B=2,D=4,T=1$  & succ & success & 6  & 8  & 2  \\
\hline
$B=2,D=4,T=5$  & succ & success & 6  & 12 & 6  \\
\hline
$B=2,D=4,T=10$ & succ & success & 6  & 17 & 11 \\
\hline
$B=2,D=4,T=20$ & succ & success & 6  & 27 & 21 \\
\hline
$B=2,D=6,T=1$  & succ & success & 8  & 10 & 2  \\
\hline
$B=2,D=6,T=5$  & succ & success & 8  & 14 & 6  \\
\hline
$B=2,D=6,T=10$ & succ & success & 8  & 19 & 11 \\
\hline
$B=2,D=6,T=20$ & succ & success & 8  & 29 & 21 \\
\hline
\end{tabular}
\end{table*}

Across all configurations, \PVaranus reaches a conclusive verdict
before the final event that makes the outcome explicit. The anticipation gain
grows linearly with the tail length $T$: it is 2 events for $T=1$, 6 events
for $T=5$, 11 events for $T=10$, and 21 events for $T=20$. This shows that,
once the execution reaches the commitment event, the practical benefit of
prediction increases with the amount of forced future behaviour that remains
before the final \texttt{mission\_complete} or \texttt{mission\_abort} event.



\subsubsection{Discussion}

Together, the rover case study and the synthetic stress test give a clear
cost--benefit picture of \PVaranus.
On the benefit side, the rover experiments show that predictive verdicts can be
obtained early enough to be operationally useful. In the nominal mission, the
coverage property is concluded eight events before mission completion; after
hazardous radiation is detected, failure of the completion and coverage
properties is detected two events before mission abort, and the abort property
is concluded one event earlier. These gains arise on traces that still conform
to the \gls{csp} model, which is precisely the target of predictive monitoring.
On the cost side, the dense synthetic benchmark shows that the predictive component is the main scalability bottleneck of the current prototype. While \Varanus remains cheap, the predictive \gls{ltl} monitor becomes expensive for larger models, both in preprocessing and per-event checking. This is a limitation of the present implementation rather than of the pipeline: the predictive monitor is modular and could be replaced by a faster automata backend, a symbolic representation, or a different predictive logic without changing the \gls{csp} conformance layer. The decision-tail benchmark nevertheless shows the benefit of this cost: once a commitment point fixes the remaining behaviour, \PVaranus returns a conclusive verdict immediately, with anticipation increasing with the forced-tail length.

Overall, the experiments show that \PVaranus preserves the conformance
guarantees of \Varanus while adding useful predictive verdicts, at the cost of
higher overhead on larger models in exchange for earlier conclusive verdicts on
longer forced continuations.

\section{Related Work}

This work lies between \emph{predictive runtime verification} and \emph{\gls{csp}-based runtime monitoring}. Model-based \gls{prv} already restricts prediction to futures admitted by a transition system, automaton, or static model \cite{DBLP:conf/nfm/ZhangLD12,DBLP:journals/jss/PinisettyJTFMP17}. \PVaranus differs in reusing the same structured \gls{csp} model both as the online conformance oracle, as in \Varanus \cite{DBLP:conf/taros/LuckcuckFF25}, and as the source of admissible futures for predictive \gls{ltl} monitoring.

Classical RV monitors finite prefixes. Giannakopoulou and Havelund compile
temporal properties into automata-based observers
\cite{DBLP:conf/kbse/GiannakopoulouH01}, while Bauer et al.\ give the standard
account of LTL and TLTL monitoring over finite prefixes
\cite{DBLP:journals/tosem/BauerLS11}. Predictive RV instead restricts
attention to model-consistent futures. Zhang et al.\ introduce predictive LTL
semantics \cite{DBLP:conf/nfm/ZhangLD12}, and Pinisetty et al.\ extend them to
timed properties \cite{DBLP:journals/jss/PinisettyJTFMP17}. These works
provide the semantic basis for our approach, but not with \gls{csp}-defined futures.

On the process-algebraic side, \gls{csp} has long been used to model ordering,
communication, and synchronisation. Tools such as FDR
\cite{DBLP:conf/tacas/Gibson-RobinsonABR14}, PAT
\cite{DBLP:conf/isola/SunLD08}, and RoboChart
\cite{DBLP:journals/sosym/MiyazawaRLCTW19} demonstrate the design-time value of
\gls{csp}-like models, but are primarily used offline. The closest runtime work is
\Varanus, which reuses a \gls{csp} specification as a runtime oracle
\cite{DBLP:conf/taros/LuckcuckFF25}. However, \Varanus is still a conformance
monitor: it detects invalid events but does not reason predictively about
future executions. Our contribution extends this model-reuse perspective with
predictive temporal reasoning.

Related robotics work has addressed RV and assurance from different angles. ROSRV and ROSMonitoring provide online monitoring frameworks for ROS \cite{DBLP:conf/rv/HuangEZMLSR14,DBLP:conf/taros/FerrandoC0AFM20}, while Desai et al.\ and SOTER combine runtime reasoning with safety assurance \cite{DBLP:conf/rv/DesaiDS17,DBLP:conf/dsn/DesaiGSST19}. Unlike these approaches, which emphasise monitoring infrastructure or safe control, our work focuses on predictive diagnosis over executions that remain conformant with a trusted behavioural model.

A final point of comparison is temporal-logic planning and synthesis for
robotics. Kress-Gazit et al.\ and Alonso-Mora et al.\ use temporal logic to
construct correct-by-design robot behaviour
\cite{DBLP:journals/trob/Kress-GazitFP09,DBLP:journals/arobots/Alonso-MoraDRRK18}.
Our contribution is complementary: rather than synthesising a controller, we
monitor a running system and predict its future outcomes under the behaviours
admitted by a \gls{csp} model.

To the best of our knowledge, no other approach combines \gls{csp}-based runtime
conformance monitoring with predictive LTL \gls{rv} in the way we
adopt. \PVaranus uses the same \gls{csp} model both as a
conformance gate for the observed event trace and as the source of admissible
futures for the predictive monitor, enabling earlier verdicts while preserving
a clear distinction between model non-conformance and temporal-property
failure.

\section{Conclusions}

We present \PVaranus, an \gls{rv} pipeline that combines \gls{csp}-based conformance checking with predictive \gls{ltl} monitoring. A single \gls{csp}
model is reused both to validate the observed event trace and to generate the
futures explored by the predictive monitor. This lets the monitor reject
invalid behaviour immediately, while classifying valid prefixes as satisfying,
violating, or inconclusive for the monitored \gls{ltl} properties.
The evaluation highlights both the benefit and the cost of this design. In the
rover case study, \PVaranus yields earlier and operationally useful verdicts by
recognising or predicting mission completion, coverage, and abort before
terminal events occur. In the synthetic benchmarks, \gls{csp} validation remains
lightweight, while the predictive monitor accounts for most of the overhead as
model size grows. The decision-tail experiments nevertheless show a clear
benefit: anticipation increases when more future behaviour is forced after the
commitment point.
Overall, \PVaranus preserves the \gls{csp}-based guarantees of \Varanus, reuses a
single design-time model throughout the monitoring pipeline, and adds useful
predictive verdicts on realistic robotic traces.

Future work includes improving predictive efficiency, enabling mitigation or
enforcement from negative verdicts, exploring richer verdict domains such as
five-valued semantics, and extending the approach to multi-component or
multi-model robotic systems.


\bibliographystyle{IEEEtran}
\bibliography{main}

\end{document}